\documentclass[sigconf,nonacm]{acmart}

\usepackage{booktabs}
\usepackage{multirow}
\usepackage{enumitem}
\usepackage{makecell}
\usepackage{amsmath}
\usepackage{graphicx}
\usepackage{xcolor}
\usepackage{balance}
\usepackage{array}
\usepackage{tabularx}
\usepackage{tikz}
\usetikzlibrary{arrows.meta,positioning,fit,backgrounds}
\newcommand{\pinsieve}{\textsc{PinSieve}}
\newcommand{\fnr}{\mathrm{FNR}}
\newcommand{\js}{\mathrm{JS}}
\newcommand{\apass}{\textsc{auto-pass}}
\newcommand{\escalate}{\textsc{escalate}}
\newcommand{\dc}{\textsc{DC-Replay}}

\title{\pinsieve: Production Selective VLM Serving and a Governed Memory Flywheel for Enterprise Content-Quality Triage}

\author{Chuqing Gao}
\email{cgao@pinterest.com}
\orcid{0009-0008-2171-5036}
\affiliation{%
  \institution{Pinterest}
  \city{New York}
  \state{NY}
  \country{USA}
}

\author{Yuanfang Song}
\authornote{Work done at Pinterest.}
\email{ysong243@wisc.edu}
\affiliation{%
  \institution{Pinterest}
  \city{Palo Alto}
  \state{CA}
  \country{USA}
}

\author{Jonathan Zhang}
\email{jonathanzhang@pinterest.com}
\affiliation{%
  \institution{Pinterest}
  \city{New York}
  \state{NY}
  \country{USA}
}

\author{Yifan Wu}
\authornotemark[1]
\email{yw693@cornell.edu}
\affiliation{%
  \institution{Pinterest}
  \city{Palo Alto}
  \state{CA}
  \country{USA}
}

\author{Vishwakarma Singh}
\authornotemark[1]
\email{vsingh014@rediffmail.com}
\affiliation{%
  \institution{Pinterest}
  \city{Palo Alto}
  \state{CA}
  \country{USA}
}

\author{Qinglong Zeng}
\email{qzeng@pinterest.com}
\affiliation{%
  \institution{Pinterest}
  \city{San Francisco}
  \state{CA}
  \country{USA}
}

\author{Andrey Gusev}
\email{andreygusev@pinterest.com}
\affiliation{%
  \institution{Pinterest}
  \city{San Francisco}
  \state{CA}
  \country{USA}
}

\begin{abstract}
Enterprise AI agents in production often need to be bounded, stateful, observable, and governable rather than fully autonomous. We present \pinsieve{}, a production case study of this design pattern in a large-scale content-quality pipeline. Its deployed component is a selective vision-language-model (VLM) Serving Agent that operates only on the grey-zone slice left unresolved by lightweight upstream models, exposes only a scalar routing score online, and preserves controlled human escalation. On this slice, the deployed system filters 2.05$\times$ more non-actionable items than the previous production module while slightly reducing estimated miss rate; after promotion, it improves review productivity by 25.7\%, reduces normalized operating cost by 16.2\%, and moves signal delivery from next-day to same-day.

We then study how to maintain such a selective model through a governed memory flywheel under selective feedback, where escalated items are reviewed by default but auto-passed items are labeled mainly through audit sampling. Feedback Memory records routing traces, observation paths, audit propensities, and replay metadata needed for evaluation and debugging. The Data Curation Agent operates as a bounded proposal-verifier loop over representative, uncertainty, recency, and fresh-review replay, with positive-rate and score-bin guardrails enforced before a batch can be accepted. In chained monthly refresh over six months of production data, this design reduces average FNR@50\% from 17.73\% under representative random replay to 13.29\%. A separate Reasoning Review Agent audits teacher-generated rationales and supports keep/repair/drop decisions, showing that supervision quality should itself be treated as part of the enterprise-agent lifecycle. Production claims in this paper are attributed only to the deployed Serving Agent; replay and rationale-review results are reported as offline or sampled-governance evidence. Beyond the focal signal analyzed here, the same serving-agent recipe has since been adopted to several additional internal signals, suggesting that the proposed method is transferable rather than a single task-specific design.
\end{abstract}


\begin{document}
\maketitle

\section{Introduction}
Large online platforms depend on early content-quality signals to route fresh content, protect downstream ranking quality, and make efficient use of expert review. In practice, however, the signals that matter most are often the ones that are hardest to obtain quickly. Lightweight front-end models and rules can handle easy cases at high throughput, but judgment-critical cases, those that depend on richer multimodal evidence or nuanced human interpretation, are often deferred to delayed features or manual review. This creates a familiar production bottleneck: the platform gets either fast but noisy decisions, or stronger decisions that arrive too late to be maximally useful.

That tradeoff is especially costly for fresh content. Engagement with new items is front-loaded, so next-day signals miss the period in which ranking and intervention are most valuable. At the same time, grey-zone traffic that cannot be confidently resolved by lightweight screening accumulates recurring human-review cost. In our setting, the operational goal is therefore not to replace the full production pipeline with a monolithic model, but to improve the boundary-case decision point under strict constraints on cost, latency, auditability, and rollback.

Recent VLMs make this tempting because they can reason directly over image-text inputs, but universal VLM serving is still too expensive and operationally blunt for this regime. Our design starts from a narrower question: can a compact VLM be inserted only where it matters most, inside an existing production cascade, while preserving human escalation and explicit threshold control? \pinsieve{} answers this with a selective Serving Agent deployed only on grey-zone traffic. The online interface is intentionally minimal: a scalar routing score that supports auto-pass versus escalate, so that the system remains easy to monitor, calibrate, and roll back. 

But deploying a selective model introduces a second systems problem: the model changes what gets observed. Escalated items are reviewed by default, whereas auto-passed items are labeled mainly through audit sampling. Naively retraining on observed feedback can therefore amplify blind spots instead of fixing them. This is where the agentic aspect of \pinsieve{} enters. We frame the post-deployment lifecycle as a bounded enterprise-agent workflow with persistent state, explicit proposal-verifier loops, structured observability, and human-controlled promotion. Feedback Memory records not only labels, but also routing traces, observation paths, audit propensities, and replay metadata. Lifecycle agents can propose replay batches, generate and review supervision, and train candidate refreshes, but they cannot directly change the online system without passing evaluation gates.

In summary, the paper supports four takeaways with separated evidence levels:
\begin{itemize}[leftmargin=*,itemsep=1pt,topsep=2pt]
    \item \textbf{Production.} A deployed selective VLM Serving Agent improves the grey-zone operating point, increasing auto-pass from 20.48\% to 41.99\% while slightly lowering estimated miss rate, and improving review productivity, cost, and latency.
    \item \textbf{Offline replay.} Under selective feedback, governed replay beats representative random replay in chained next-window evaluation, reducing average FNR@50\% from 17.73\% to 13.29\%.
    \item \textbf{Observability and governed supervision.} Feedback Memory records routing traces, observation paths, audit propensities, and replay metadata so that replay, evaluation, debugging, and promotion are grounded in what the system actually observed. In the long-running flywheel, teacher-generated rationales are not recycled blindly and undergone review before re-entering supervision.
    \item \textbf{Transfer evidence.} The same serving-agent recipe has since been adopted to several additional internal signals, suggesting that the transferable artifact is the bounded serving contract rather than a single task-specific threshold.
\end{itemize}

\section{Related Work and Positioning}
\textbf{Enterprise agentic systems and bounded automation.} Recent agent work emphasizes reasoning, tool use, orchestration, and persistent state~\cite{wang2024survey,yao2022react,wu2023autogen}, while industrial human-in-the-loop systems emphasize bounded automation, escalation, and auditability in production review workflows~\cite{tanaka2020security,ramachandran2020menu,grbovic2022airbnb,markov2023undesired}. \pinsieve{} is closest to the latter regime: unlike general assistant-style agents, the online actor here has a deliberately narrow action space and operates inside an existing production cascade. Our contribution is not open-ended online autonomy, but a governed lifecycle.

\textbf{Multimodal serving, distillation, and rationale supervision.} Prior multimodal work established strong image--text representation learning and instruction-following backbones for classification and understanding~\cite{lu2019vilbert,li2019visualbert,radford2021clip,liu2023llava,wang2024qwen2vl}. Separate lines of work study compact students and richer supervision through knowledge distillation, chain-of-thought distillation, and rationale-guided training~\cite{hinton2015distilling,sanh2019distilbert,jiao2020tinybert,hsieh2023stepbystep,dipalo2024pgkd,agrawal2025rationale}. Our setting is narrower and more deployment-constrained than these prior lines of work. While we use teacher-generated rationales for distillation, we treat them as a governed lifecycle artifact rather than a one-shot training signal. For these reasons, deployment-oriented industrial multimodal systems are more relevant to our setting than raw VLM capability comparisons~\cite{zhang2024toxic,shi2024cpfd,dong2024coefvq}. In addition, we are not trying to maximize general multimodal capability; instead, we focus on grey-zone triage, where lightweight upstream components cannot confidently resolve.

\textbf{Selective feedback, replay, and continual adaptation.} Active learning and continual-learning methods study informative querying, replay, and adaptation under evolving distributions~\cite{settles2009active,li2024deepactive,ash2020badge,cacciarelli2024streams,zliobaite2014stream,gama2014conceptdrift,hinder2024drift,parisi2019continual,rolnick2019replay,zhou2024ptmcl}. The key difference in \pinsieve{} is selective observability: labels are not obtained by freely querying an unlabeled pool, but are revealed through escalation and limited audits after deployment. 

\section{System Design}
\subsection{Architecture Overview}
\begin{figure*}[t]  
  \centering
  \includegraphics[width=\textwidth]{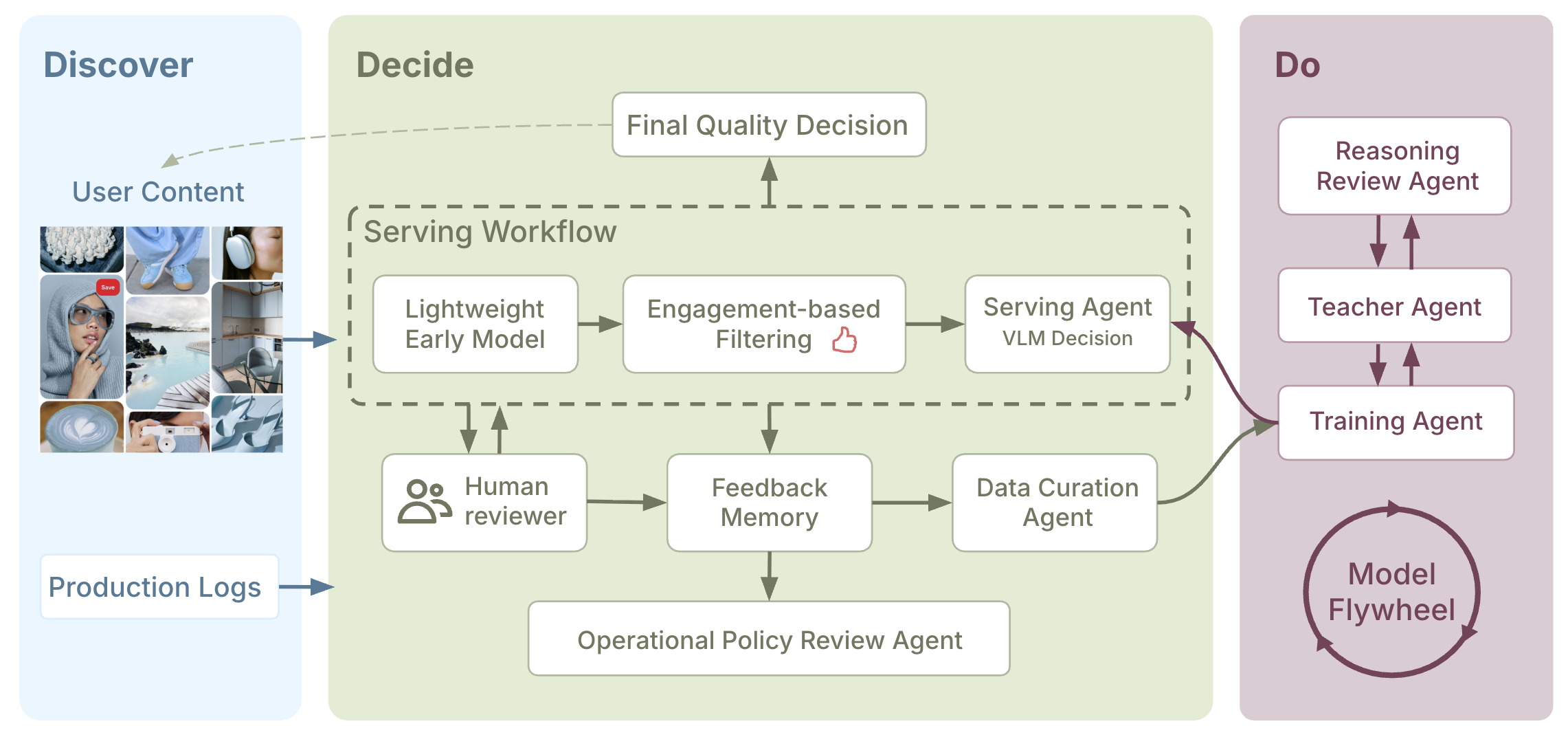} 
  \caption{\pinsieve{} System Architecture}
  \label{fig:overview}
\end{figure*}

Figure~\ref{fig:overview} summarizes \pinsieve{} as a governed enterprise-agent lifecycle. The design separates online action from offline improvement. The deployed Serving Agent performs selective triage. The Teacher Agent creates rationale-based supervision. The Reasoning Review Agent audits rationales and issues keep/repair/drop decisions. Feedback Memory stores versioned serving traces, labels, audit metadata, and replay metadata. The Data Curation Agent builds replay batches under guardrails. The Training Agent creates candidate checkpoints, and the Model Registry supports staged validation and rollback. Table~\ref{tab:contracts} makes the workflow contract explicit: each agent reads a restricted slice of state, produces a narrow action, and is bounded by verification or human sign-off. In other words, the paper's agentic claim is not that the system plans freely, but that it runs a stateful, instrumented, and governable loop over serving, memory, curation, review, and promotion.

\begin{table}[t]
\centering
\caption{Bounded enterprise-agent contracts in \pinsieve{}. The online agent is deployed; lifecycle agents are offline or shadow-validated.}
\label{tab:contracts}
\small
\begin{tabularx}{\columnwidth}{p{0.18\columnwidth}p{0.25\columnwidth}p{0.18\columnwidth}X}
\toprule
Agent & Reads / state & Action & Guardrail \\
\midrule
Serving & Grey-zone item, threshold, model version & Emit routing score & Scalar-only interface \\
\hline
Teacher & Labeled item, stable guidance & Generate rationale target & Cannot overwrite human label \\
\hline
Reasoning Review & Image, rationale, label & Keep / repair / drop verdict & Cannot redefine policy \\
\hline
\makecell[l]{Data \\ Curation}& Memory summaries, slice stats, replay budget & Propose replay plan & Verified against rate/divergence constraints \\
\hline
Training & Curated batch, previous checkpoint &\makecell[l]{Train \\ Candidate \\ checkpoint} & Shadow validation needed \\
\hline
Operational Review & Audit false negatives & Root-cause summary & Diagnostic only \\
\bottomrule
\end{tabularx}
\end{table}

\subsection{Serving Agent: Selective VLM Triage}
The Serving Agent is the only online actor in \pinsieve{}. It is inserted after lightweight upstream components resolve easy cases, receives the item image and associated text context, and emits a score $s_\theta(x)$ for the target low-quality class. Given a threshold $\tau$, the online routing rule is
\begin{equation}
    a(x)=\begin{cases}
    \apass, & s_\theta(x)<\tau,\\
    \escalate, & s_\theta(x)\ge \tau .
    \end{cases}
\end{equation}
The threshold is selected to satisfy a target operating constraint on false negative rate. Auto-passed items remain subject to audit sampling, and escalated items remain in the human-in-the-loop path. This preserves operational control while reducing unnecessary review load.

At inference time, the Serving Agent exposes only a scalar score. Although the model is trained with structured decision-rationale targets, rationale text is not used as an online explanation or decision artifact. Let $z_{\mathrm{no}}$ and $z_{\mathrm{yes}}$ be logits for negative and positive decision tokens at a fixed position. The deployed score is
\begin{equation}
    s_\theta(x)=\frac{\exp(z_{\mathrm{yes}})}{\exp(z_{\mathrm{no}})+\exp(z_{\mathrm{yes}})}.
\end{equation}
This interface keeps the production action legible, cheap, auditable, and easy to monitor. The deliberately narrow action space is a deployment choice: the system gains stronger multimodal judgment while preserving threshold control, rollback, and human escalation.

\subsection{Teacher and Reasoning Review Agents}
The Teacher Agent is used offline to generate structured supervision for the compact student. Its purpose is not to update quality guidance. Instead, it converts stable signal definitions, reviewer-aligned outcomes, and item context into reusable decision-rationale targets. For a labeled item $(x_i,y_i)$, the supervision target is serialized as
\begin{quote}\small
\texttt{\{decision\}, \{routing outcome\}. \{grounded rationale\}}
\end{quote}
where the rationale must be consistent with the reviewer outcome, reference only observable evidence, and avoid exposing internal guidance. This allows the student to learn a compact approximation of the decision process without prompt-time access to long guidelines.This allows the student to learn a compact approximation of the decision process without prompt-time access to long guidelines.

The Reasoning Review Agent audits this supervision before it enters future training. Given image/context $x$, teacher rationale $r$, and human decision $y$, it assigns a review verdict such as faithful, faithful-but-wrong-conclusion, confabulation, forced rationalization, spurious feature focus, or over-interpretation. These categories are mapped to \textsc{keep}, \textsc{repair}, or \textsc{drop}. Repair cases are returned to the Teacher Agent with reviewer feedback; drop cases are excluded from rationale-supervised training.

\subsection{Feedback Memory}
Feedback Memory is the observability layer of the lifecycle. Each item records four groups of fields: serving trace (model version, score, threshold, route, time bucket), observation path (reviewer or audit outcome, label source, audit probability), lightweight context summaries, and replay metadata such as score bin or uncertainty flag.

The observation path is essential because escalated items and audited auto-passed items are not exchangeable samples. Recording whether a label came from escalation or audit supports inverse-propensity-weighted evaluation, lets the Data Curation Agent reason about under-observed regions, and gives operators enough traceability to debug the flywheel.

\subsection{Data Curation Agent: Governed Replay}
At refresh step $t$, the Data Curation Agent observes cumulative memory $\mathcal{M}_t$, previous checkpoint $\theta_{t-1}$, estimated natural positive rate $\pi_{\mathrm{nat},t}$, and reference score-bin distribution $p_{\mathrm{ref},t}$. It does not directly choose a new model; instead, it proposes a replay plan against persistent state and verifier feedback, then refines that proposal until it satisfies deployment-facing constraints. Concretely, it proposes a fixed-size replay batch
\begin{equation}
B_t = B^{\mathrm{rep}}_t \cup B^{\mathrm{unc}}_t \cup B^{\mathrm{recent}}_t,
\end{equation}
where representative replay anchors the natural mixture, uncertainty replay emphasizes boundary cases, recent replay tracks newer traffic.

The proposal is accepted only if it satisfies deployment-facing guardrails:
\begin{equation}
    |\pi(B_t) - \pi_{\mathrm{nat},t}| \le \delta_{\pi}, \quad
    \js\big(p_{\mathrm{bin}}(B_t) \| p_{\mathrm{ref},t}\big) \le \delta_{\js}.
\end{equation}
If a batch violates these constraints, the agent increases the representative fraction and shrinks targeted sources proportionally. In this sense, \dc{} is adaptive but bounded: it can pursue targeted adaptation only when the resulting batch remains close to the serving-facing distribution.

\subsection{Training Agent and Refresh Protocol}
The Training Agent fine-tunes candidate Serving Agents on curated batches with a composite supervised objective that combines sequence targets, explicit decision-token classification, and an optional separation term to widen the score gap between positives and negatives. For chained refresh, candidate $\theta_t$ is trained from $\theta_{t-1}$ on $B_t$ and evaluated on the next unseen monthly window. Promotion is blocked unless the candidate improves target metrics under fixed safety guardrails and passes shadow testing. Operationally, refresh follows a bounded loop: candidates are trained on curated supervision, evaluated on reviewed and audited slices, shadow-tested without live action, and promoted only if they improve target metrics under fixed safety guardrails.

\subsection{Operational Review}
The Operational Review module analyzes audit-sampled false negatives after serving. Rather than changing online decisions directly, it assigns structured root-cause summaries such as score calibration error, model perception error, or policy ambiguity. These summaries are used as diagnostic evidence for later debugging, policy review, and future curation.

\paragraph{Design constraints and operational controls.}
\pinsieve{} is shaped by both operating constraints (e.g., selective feedback and the cost of universal VLM serving) and lifecycle risks (e.g., replay distortion and silent refresh regression). Table~\ref{tab:guardrails} summarizes the corresponding design responses and controls.

\begin{table}[t]
\centering
\caption{Design constraints and operational controls in \pinsieve{}.}
\label{tab:guardrails}
\small
\begin{tabularx}{\columnwidth}{p{0.38\columnwidth}X}
\toprule
Constraint or risk & Design response / control \\
\midrule
Universal VLM serving is too costly
& Invoke the VLM only on grey-zone traffic. \\

Labels are observed selectively
& Record escalation vs.\ audit path and audit probability in Feedback Memory. \\

Teacher rationales can be fluent but unsupported
& Apply keep/repair/drop review before replay into supervision. \\

Candidate refresh can regress silently
& Require next-window evaluation, shadow testing, and rollback gates. \\

Drift may remain hidden in aggregate metrics
& Summarize audit failures into root-cause categories for diagnosis. \\
\bottomrule
\end{tabularx}
\end{table}

\section{Evaluation}
For a quick read, Tables~\ref{tab:prod} and~\ref{tab:refresh} contain the two main empirical messages of the paper. Table~\ref{tab:prod} reports the only \textbf{production} results and shows that the deployed Serving Agent improves auto-pass capacity, review productivity, operating cost, and signal latency on the grey-zone slice. Table~\ref{tab:refresh} reports the main \textbf{offline replay} result and shows that governed replay improves next-window serving-facing risk relative to representative random replay under the same selective-feedback protocol. The remaining experiments explain how the deployed model was trained, why selective-feedback evaluation needs explicit observability over audit paths and propensities, and why the replay and governance claims should be interpreted separately from production impact.

\subsection{Setup and Metrics}
We evaluate one representative binary content-quality signal with consistent expert labels and direct production relevance. The main offline experiments use six months of production data. Each item contains an image and associated text context, and the positive class denotes the target low-quality condition for the signal.

Because \pinsieve{} is a selective triage layer, we emphasize operating-point metrics. $\fnr@X\%$ is the false negative rate when the bottom $X\%$ of examples by score are auto-passed. Unless stated otherwise, we use inverse propensity weighting (IPW) to correct for audit sampling in the auto-pass region. We also report PR-AUC for ranking quality, review productivity, normalized operating cost, and signal delivery latency.

\subsection{Evaluation Questions}
We organize the empirical study around four questions that match the lifecycle decomposition. \textbf{Q1:} Does the deployed Serving Agent improve the production operating point in the grey-zone slice? \textbf{Q2:} Which supervised recipe produces the static model used online? \textbf{Q3:} Can feedback memory refresh improve future windows without destabilizing selective serving? \textbf{Q4:} Are rationale review and operational failure review useful governance mechanisms, even before they are credited with online gains? Table~\ref{tab:evalplan} maps these questions to evidence.

\begin{table}[t]
\centering
\caption{Evaluation plan. Production, offline replay, and sampled-governance claims are separated to avoid over-attribution.}
\label{tab:evalplan}
\small
\begin{tabularx}{\columnwidth}{p{0.12\columnwidth}p{0.30\columnwidth}X}
\toprule
Q & Component & Evidence \\
\midrule
Q1 & Serving Agent & Shadow test and production metrics: FNR, auto-pass, cost, ETA \\
Q2 & Training recipe & Static ablations over 2B open-source VLM with/without KD and rationale response \\
Q3 & Data Curation Agent & Chained monthly refresh with next-window evaluation and IPW metrics \\
Q4 & Review agents & Sampled rationale audit and audit-failure root-cause summaries \\
\bottomrule
\end{tabularx}
\end{table}

\subsection{Attribution and Observation Paths}
We separate results into \textbf{Production}, \textbf{Offline replay}, and \textbf{Sampled governance} to avoid over-attribution: only production results support direct online claims. Under selective feedback, escalated items are usually labeled, auto-passed items are only partially observed through audit, and many auto-passed items remain unlabeled; this distinction matters for both IPW-based evaluation and replay design.

\subsection{Selective-Feedback Estimation}
Because labels in the auto-pass region are only partially observed through audit, we report miss rate with inverse-propensity weighting rather than treating observed labels as representative of all served traffic. For the bottom-$X\%$ auto-pass set $A_X$, with audit probability $q_i$ and weight $w_i=1/q_i$, we estimate
\begin{equation}
\widehat{\fnr}_{\mathrm{IPW}}(X)=
\frac{\sum_{i\in A_X \cap O} w_i\mathbf{1}[y_i=1]}
{\sum_{i\in O} w_i\mathbf{1}[y_i=1]},
\end{equation}
where $O$ denotes examples whose labels are observed through review or audit. In practice, we use logged audit probabilities, clip extreme weights, and report unweighted observed-population metrics as a secondary sanity check. For refresh experiments, thresholds are selected on the current window and then frozen for the next unseen monthly window.

\subsection{Production Impact of the Deployed Serving Agent}
The selective VLM layer was first shadow-deployed for three weeks alongside the previous production module. After dry-run validation, it was promoted to production. Table~\ref{tab:prod} shows the observed impact. On the grey-zone slice, \pinsieve{} filters 2.05$\times$ more non-actionable items while slightly reducing estimated miss rate. At the operational level, it improves review productivity by 25.7\%, lowers normalized cost from 1.000 to 0.838, and moves final signal delivery from next-day to same-day. These are the only metrics in the paper that support direct online-impact claims, because they are measured from the deployed serving path rather than from offline replay or governance analysis.

\begin{table}[t]
\centering
\caption{[Production] Impact of the deployed selective VLM serving layer. The key takeaway is 2.05$\times$ higher auto-pass capacity on the grey-zone workload with lower normalized cost and same-day delivery.}
\label{tab:prod}
\small
\begin{tabular}{lcc}
\toprule
Model & False negative rate & Auto-pass rate \\
\midrule
Previous production module & 13.62\% & 20.48\% \\
\pinsieve{} & 13.18\% & 41.99\% \\
\bottomrule
\end{tabular}
\vspace{4pt}

\begin{tabular}{lccc}
\toprule
Model & Review productivity & Cost & ETA \\
\midrule
Previous pipeline & 1.205 & 1.000 & Next-day \\
\pinsieve{} & 1.515 & 0.838 & Same-day \\
\bottomrule
\end{tabular}
\end{table}

\subsection{Static Model Training}
Table~\ref{tab:static} summarizes the \emph{offline static-training} ablations that determine the deployed recipe. All fine-tuned models use the same 2B open-source VLM backbone and 200K training samples. Thresholds are selected so that FNR is closest to but does not exceed 10\%.

Fine-tuning is already highly effective: the strongest non-distilled model increases auto-pass rate from 10.77\% to 38.85\% at similar FNR. Knowledge distillation and rationale-conditioned responses further improve the operating frontier. The final recipe reaches 46.90\% auto-pass at 8.71\% FNR. Rationale distillation also moves long guidance from prompt-time injection to training-time supervision, reducing prompt overhead during both training and inference.

\begin{table}[t]
\centering
\caption{[Offline static training] Model ablations used to choose the deployed Serving Agent recipe.}
\label{tab:static}
\small
\begin{tabular}{lcc}
\toprule
Model & FNR & Auto-pass rate \\
\midrule
Base model & 9.74\% & 10.77\% \\
Best fine-tune without KD & 9.78\% & 38.85\% \\
KD + sequence loss & 9.30\% & 40.32\% \\
KD + composite loss & 9.59\% & 42.93\% \\
KD + composite loss + rationale response & \textbf{8.71\%} & \textbf{46.90\%} \\
\bottomrule
\end{tabular}
\end{table}

We also evaluated preference-style optimization, including many DPO-style variants, on top of supervised fine-tuning. In our setting, these methods provided negligible gains and were sometimes unstable, likely because the deployed routing interface depends on a single decision-token score, whereas preference objectives optimize relative response likelihood at the sequence level. We therefore retain supervised training as the main optimization recipe.

\subsection{Refresh from Feedback Memory}
We simulate chained monthly refresh over six consecutive production windows. At each step, the Data Curation Agent selects a 50K-example refresh batch from Feedback Memory, trains a candidate model, and evaluates it on the next unseen monthly window. This forward evaluation matches the enterprise maintenance problem: the goal is not to improve on already observed feedback, but to keep the deployed model stable as traffic changes.

A stress test shows why governance is necessary. Representative random replay is competitive with uncertainty replay and safer than aggressive hard-error replay. Model-error replay collapses under the selective-feedback setting, indicating that naive mining can overfit distorted observed labels rather than improving serving-facing behavior.

\begin{table}[t]
\centering
\caption{[Offline replay] Stress test on the final monthly window. Aggressive hard-error replay is unstable under selective feedback.}
\label{tab:stress}
\small
\begin{tabular}{lccc}
\toprule
Replay strategy & PR-AUC & FNR@30\% & FNR@50\% \\
\midrule
Uncertainty & 0.589 & 6.99 & 16.00 \\
Random & 0.585 & 7.57 & 15.56 \\
Hybrid & 0.533 & 6.24 & 21.35 \\
Model-error & 0.072 & 60.72 & 71.92 \\
\bottomrule
\end{tabular}
\end{table}

Table~\ref{tab:refresh} compares the main \emph{offline replay} choices averaged across cycles. Random replay is a strong operational baseline, improving PR-AUC and reducing FNR@50\% relative to no refresh. \dc{} further improves the metrics most aligned with selective triage, reducing average FNR@50\% from 17.73\% to 13.29\%.

\begin{table}[t]
\centering
\caption{[Offline replay, IPW] Chained monthly refresh under serving-distribution evaluation. The key takeaway is a reduction in average IPW-FNR@50\% from 17.73\% under representative random replay to 13.29\% under \dc{}.}
\label{tab:refresh}
\small
\begin{tabular}{lcccc}
\toprule
Method & PR-AUC & FNR@30\% & FNR@40\% & FNR@50\% \\
\midrule
Static Serving Agent & 0.4784 & 13.00\% & 17.06\% & 21.69\% \\
Random replay & 0.5310 & 8.59\% & 13.03\% & 17.73\% \\
\dc{} & \textbf{0.5542} & \textbf{5.95\%} & \textbf{9.54\%} & \textbf{13.29\%} \\
\bottomrule
\end{tabular}
\end{table}

\subsection{Supervision and Operational Review}
The Reasoning Review Agent is evaluated through sampled spot checks of teacher-generated rationales. In the sample set, $92.2\%$ were judged faithful and $7.8\%$ were flagged as low-faithfulness. Among the flagged cases, $44.4\%$ appeared to be annotation noise and were marked for removal from future supervision, while $38.8\%$ were generation errors that could be repaired by regenerating the rationale with reviewer feedback. We report this as a \emph{sampled governance study} rather than an accuracy gain: it shows that supervision quality is observable and governable.

The operational review module analyzes false negatives from the audit stream. It assigns root-cause categories such as score calibration error, model perception error, and policy ambiguity. This module does not change online decisions. Its value is to transform difficult audit failures into structured evidence for future curation, debugging, and policy review. In an enterprise setting, this type of structured post-hoc analysis is often more actionable than a single aggregate miss-rate number because it distinguishes whether the next intervention should target thresholding, model capacity, replay coverage, or policy clarification.

\section{Discussion and Lessons}
Selective deployment can be more valuable than universal autonomy. The production gain comes from applying stronger VLM reasoning to a narrow high-value slice, not from replacing the whole cascade. Observability is part of the agent design. Feedback memory, audit propensities, and root-cause summaries are what make selective replay and staged promotion governable. Feedback memory must record how labels are observed. Without escalation/audit provenance and audit probabilities, both evaluation and replay can become biased.

Representative random replay is a strong baseline because it preserves the serving-facing mixture; targeted replay should be added only under explicit guardrails. Rationale supervision is useful but must be governed: fluent rationales are not always faithful, so keep/repair/drop review reduces the risk of amplifying supervision noise. More broadly, agentic control should match decision complexity. In simple single-signal refresh loops, deterministic policies can be difficult to beat; LLM control is more useful when decisions are multi-objective and semantically coupled.

The detailed experiments in this paper focus on one representative signal because that is the only setting for which we currently have a fully aligned production, replay, and governance analysis. However, the Serving Agent recipe has since been adopted to several additional internal signals. We therefore view the deployment pattern for judgment-critical triage rather as transferable. This pattern is most useful when a lightweight front-end already resolves easy cases, the target signal is judgment-critical, at least some audit or delayed-feedback channel exists, and operators are willing to enforce explicit rollout gates.

\section{Scope and Next Steps}
This workshop paper focuses on one representative content-quality signal and uses it to study a bounded enterprise-agent workflow for selective serving and governed post-deployment adaptation. The same serving-agent recipe has since been adopted to several additional internal signals, suggesting that the serving contract is transferable beyond the focal signal studied here. The current Data Curation Agent uses bounded proposal-and-verification logic, and richer orchestration is a natural extension as the control problem broadens. The supervision review study likewise motivates larger-scale measurement of downstream effects from rationale filtering. Some implementation details, internal guidelines, and example cases are intentionally abstracted to satisfy compliance requirements.

\section{Conclusion}
\pinsieve{} pairs a deployed selective VLM Serving Agent with a governed lifecycle for post-deployment maintenance. In production, the Serving Agent improves the grey-zone operating point by increasing auto-pass from 20.48\% to 41.99\%, improving review productivity from 1.205 to 1.515, reducing normalized operating cost from 1.000 to 0.838, and moving signal delivery from next-day to same-day. Offline, we show that safe refresh under selective feedback requires governed replay rather than naive retraining: average FNR@50\% drops from 17.73\% with representative random replay to 13.29\% with DC-Replay. Finally, sampled review shows that teacher-generated rationales should not be recycled blindly in a long-running flywheel, but should undergo explicit keep/repair/drop review before re-entering supervision. Taken together, these results support a bounded deployment recipe: online gains come from selective serving, while replay, supervision reuse, and refresh remain governed offline.

\begin{acks}
We would like to thank the GenAI serving team for help with the online deployment, thank the Ops team for their cooperation, and Manisha Srivastava,Siddhartha Reddy Jonnalagadda, Faisal Farooq and Balaji Krishnapuram for their feedback and support throughout the project.
\end{acks}


\balance
\bibliographystyle{ACM-Reference-Format}
\bibliography{paper2}
\end{document}